\documentclass{article} 
\usepackage[final]{colm2026_conference}

\usepackage{microtype}
\usepackage{hyperref}
\usepackage{url}
\usepackage{booktabs}

\usepackage{lineno}

\definecolor{darkblue}{rgb}{0, 0, 0.5}
\hypersetup{colorlinks=true, citecolor=darkblue, linkcolor=darkblue, urlcolor=darkblue}

\title{\textit{Growing Pains}: Extensible and Efficient LLM Benchmarking Via Fixed Parameter Calibration}

\newcommand{\HUJI}{\textsuperscript{1}}
\newcommand{\IBM}{\textsuperscript{2}}
\newcommand{\MITaff}{\textsuperscript{3}}
\newcommand{\Technion}{\textsuperscript{4}}
\newcommand{\AllenAI}{\textsuperscript{5}}
\newcommand{\authorsep}{\hspace{1.8em}}

\author{
\vspace{-0.8em}\\
\textbf{Eliya Habba}\HUJI \authorsep
\textbf{Itay Itzhak}\textsuperscript{1,4} \authorsep
\textbf{Asaf Yehudai}\IBM \authorsep
\textbf{Yotam Perlitz}\IBM \authorsep
\textbf{Elron Bandel}\IBM
\\[2.3ex]
\textbf{Michal Shmueli-Scheuer}\textsuperscript{2}\thanks{Equal contribution.} \authorsep
\textbf{Leshem Choshen}\textsuperscript{2,3}\footnotemark[1] \authorsep
\textbf{Gabriel Stanovsky}\textsuperscript{1,5}\footnotemark[1]
\\[0.8ex]
\HUJI The Hebrew University of Jerusalem \authorsep
\IBM IBM Research \authorsep
\MITaff MIT
\\
\Technion Technion \authorsep
\AllenAI Allen Institute for AI
\\[0.8ex]
\texttt{eliya.habba@mail.huji.ac.il}
}
\usepackage{subcaption}
\usepackage{makecell}
\usepackage{amsmath}
\usepackage{multirow}
\usepackage{graphicx}
\usepackage{xcolor}
\usepackage{amssymb}

\DeclareMathOperator*{\argmax}{arg\,max}

\usepackage[normalem]{ulem}
\usepackage{pdfcomment}
\usepackage{todonotes}
\usepackage{cooltooltips}

\newcommand{\com}[1]{}

\newcommand{\resolved}[1]{}

\begin{document}

\ifcolmsubmission
\linenumbers
\fi

\maketitle

\begin{abstract}
The rapid release of both language models and benchmarks makes it increasingly costly to evaluate \emph{every} model on \emph{every} dataset. In practice, models are often evaluated on different samples, making scores difficult to compare across studies. To address this, we propose a framework based on multidimensional Item Response Theory (IRT) that uses anchor items to calibrate new benchmarks to the evaluation suite while holding previously calibrated item parameters fixed.
Our approach supports a realistic evaluation setting in which datasets are introduced over time and models are evaluated only on the datasets available at the time of evaluation, while a fixed anchor set for each dataset is used so that results from different evaluation periods can be compared directly. In large-scale experiments on more than $400$ models, our framework predicts full-evaluation performance within 2--3 percentage points using only $100$ anchor questions per dataset, with Spearman $\rho \geq 0.9$ for ranking preservation. This shows that benchmark suites can grow over time while preserving score comparability, since adding a new dataset requires running existing models only on that dataset’s anchors.~\footnote{Code and data are available at: \url{https://eliyahabba.github.io/growing-pains/}.}
\end{abstract}

\section{Introduction}
\label{sec:intro}


The growing number of LLMs and evaluation benchmarks makes exhaustive evaluation of every model on every sample prohibitively expensive~\citep{hofmann2025fluidlanguagemodelbenchmarking, kiela2021dynabenchrethinkingbenchmarkingnlp,perlitz2024efficientbenchmarkinglanguagemodels}. Instead,  existing leaderboards opt to evaluate models against different subsets of evaluation instances.
As a consequence, the public record of results is fragmented, with test suites overlapping only partially. As a result, both absolute results and relative rankings can vary significantly with benchmark selection~\citep{perlitz2024llmbenchmarksagreefixing}. 


\begin{figure*}[tb!]
    \centering
    \includegraphics[width=\textwidth]{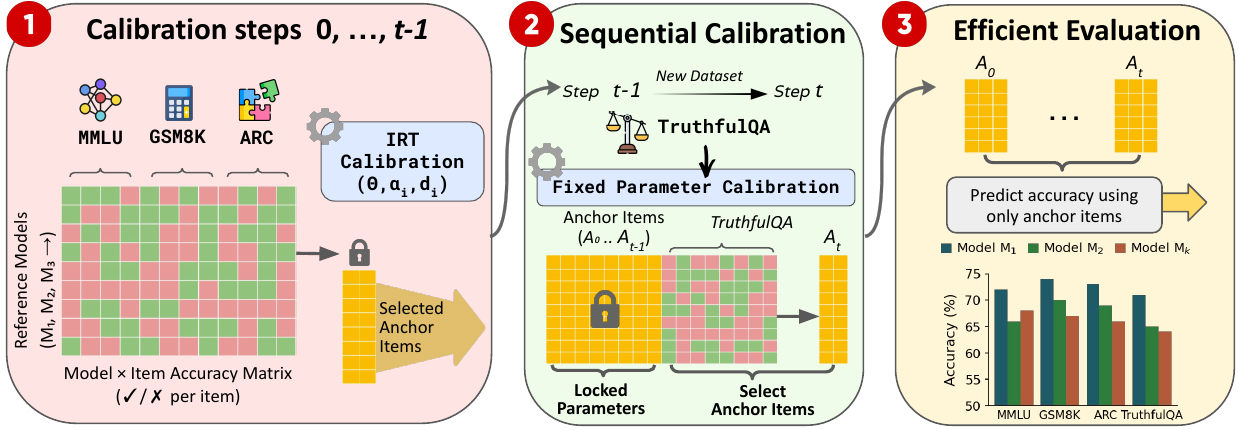}
    \caption{Fixed parameter calibration enables extensible evaluation as new benchmarks are added over time. (1) Base datasets (e.g., MMLU, GSM8K) are calibrated jointly on reference models to define initial anchor item parameters. (2) At each subsequent step t, a new dataset is integrated by estimating its item parameters while holding all previously calibrated anchor parameters fixed (locked icons). (3) Once calibrated, the accumulated anchors serve as a compact proxy for the full suite, enabling performance prediction from anchor responses alone.}
    \label{fig:summary_dashboard}
\end{figure*}

In this work, we propose an efficient and reproducible real-world evaluation framework. Specifically, we consider a setting in which datasets are released over time, and models are evaluated on the datasets available at the time of evaluation. 
As illustrated in Figure~\ref{fig:summary_dashboard}, we introduce a psychometric framework based on multidimensional item response theory (IRT) and a small set of anchor questions \citep{Kolen2004TestES}. 
We start by calibrating an IRT model on a small set of datasets. When new datasets are added, we update the evaluation pipeline via fixed parameter calibration. In this process, previously learned item parameters are held fixed, and only the parameters of newly introduced items are estimated using reference models that connect old and new items. This allows the benchmark suite to grow while preserving the existing calibration.

This setup supports two use cases. For new-model evaluation, we test only the anchor questions and use the calibrated IRT model to infer performance over the larger historical pool, reducing the amount of inference required. For new-dataset integration, we estimate its item parameters from the reference models' responses while keeping the existing anchor parameters fixed. We then run the existing models only on the anchors selected from the new dataset and use these responses to estimate their performance on the full dataset.

We evaluate our approach on two large-scale benchmark suites: the Open LLM Leaderboard~\citep{fourrier2024open} (6 datasets, 395 models) and MMLU~\citep{hendrycks2021measuringmassivemultitasklanguage} (57 subject subsets, 428 models). Our experiments simulate the sequential addition of new benchmarks and the arrival of new models over time, and compare fixed parameter calibration against concurrent re-calibration and random sampling.

Across both suites, fixed parameter calibration achieves stable accuracy as the benchmark pool expands. With only 100 anchor questions per dataset, prediction error is typically low (around 2–3\%), and it does not accumulate as more datasets are added. A cost analysis further shows that concurrent calibration reselects anchors across the accumulated suite at every step, requiring models to be evaluated on an updated anchor set that grows with the suite. Its cost therefore grows with the suite. Fixed parameter calibration retains the existing anchors and adds only anchors from the newly added dataset. For previously evaluated models, the cost of each new dataset therefore remains constant. We also observe diminishing returns from very large anchor sets and that a moderate number of reference models suffices for accurate prediction as the benchmark suite grows. Together, these results address both use cases: evaluating new models from anchor responses alone and integrating new datasets while retaining the existing anchors.


To conclude, our contributions are: (1) we formulate LLM evaluation under evolving dataset coverage as a scale-linking problem, where datasets are introduced over time and models are evaluated only on the datasets available at the time of evaluation; (2) we introduce a multidimensional IRT framework with fixed anchor sets and sequential fixed-parameter calibration so that results collected at different times can be compared directly; and (3) we show on suites derived from the Open LLM Leaderboard and MMLU that this approach provides a cost-effective approximation to full evaluation while largely preserving model rankings.


\section{Background}
\label{sec:background}
This section introduces the psychometric concepts underlying our framework:
Item Response Theory (IRT) as the measurement model, the limitations of
concurrent calibration under evolving benchmarks, and fixed parameter
calibration as the mechanism for integrating new benchmarks without recalibrating the existing suite.

\subsection{Item Response Theory}
\label{sec:irt}

Item Response Theory (IRT) is a probabilistic measurement framework that models each response as a function of latent model abilities and individual item characteristics~\citep{Lord1984ComparisonOI, polo2024tinybenchmarksevaluatingllmsfewer}.
Whereas classical test theory characterizes performance through aggregate
scores, IRT estimates separate parameters for each item and each model,
enabling fine-grained analysis of what drives correctness.

We employ the Multidimensional 2-Parameter Logistic  variant of
IRT (MIRT 2PL), which allows
each item to load on multiple latent dimensions, capturing distinct
skills required in different questions and facilitating cross-dataset linking. A model's
ability is represented as a vector $\boldsymbol{\theta}$ over these
dimensions, and the probability of a correct response for item~$i$ is:

\begin{equation}
\label{equation_1}
P(y_i=1 \mid \boldsymbol{\theta}) = \frac{\exp(\mathbf{a}_i^T \boldsymbol{\theta} + d_i)}{1 + \exp(\mathbf{a}_i^T \boldsymbol{\theta} + d_i)}
\end{equation}

\noindent where $\mathbf{a}_i$ is the discrimination vector, representing
the item's sensitivity to each latent dimension, and $d_i$ is an intercept
term related to item difficulty.

\subsection{IRT under evolving benchmarks}
\label{sec:irt_evolving}

IRT-based methods have been shown to enable efficient LLM evaluation by estimating performance from small, representative item subsets~\citep{polo2024tinybenchmarksevaluatingllmsfewer}. These methods rely on \textit{concurrent calibration}, which jointly estimates all item parameters ($\mathbf{a}_i$, $d_i$) and model abilities ($\boldsymbol{\theta}$) over a fixed benchmark~\citep{Lord1984ComparisonOI}. While effective for static evaluations, concurrent calibration becomes problematic when benchmarks evolve. Each new benchmark requires recalibrating the accumulated suite and reselecting anchors across the previously added datasets. Models must therefore be evaluated on an updated anchor set that grows as the suite expands. Moreover, re-estimating all parameters jointly shifts the calibration. Items originally calibrated with parameters $\mathbf{a}_i$,
$d_i$ receive updated estimates $\mathbf{a}_i'$, $d_i'$, rendering historical ability estimates $\boldsymbol{\theta}$ incomparable. 
We therefore need a calibration strategy that preserves the existing scale as new benchmarks are added, without requiring each model to be evaluated on an increasingly large anchor set as the suite grows.

\subsection{Test equating and fixed parameter calibration}
\label{sec:fpc}

Test equating addresses precisely this need. In psychometric practice, test equating makes scores from different test forms comparable~\citep{Kolen2004TestES}, typically through a set of shared anchor items that serve as a common reference. When new items are calibrated to an existing evaluation suite by holding anchor item parameters constant, the procedure is known as
fixed parameter calibration~(FPC;~\citealp{kim1998comparison}).

Concretely, when a new dataset is added, FPC holds anchor item parameters
($a_i$, $d_i$ for items in the anchor set) fixed and estimates only the parameters of newly introduced items. Because anchor parameters are never
modified, model abilities $\boldsymbol{\theta}$ retain consistent meaning across calibration steps, keeping scores comparable over time.
Fixed parameter calibration therefore supports efficient evaluation of new models from the accumulated anchors. For an evolving benchmark suite, it enables new datasets to be integrated while retaining the existing anchors across calibration steps. While FPC is well established in psychometrics, it has not been applied to LLM evaluation. In Section~\ref{sec:methods}, we apply this procedure sequentially as new datasets are added to a growing benchmark suite, so that the scores remain comparable as new datasets accumulate.


\section{Method: sequential fixed parameter calibration}
\label{sec:methods}
To achieve extensible evaluation, we consider a sequence of dataset releases in different time steps t = 0, 1, \ldots .
At each time step, we seek to find a set of representative anchors for the new dataset, which will remain fixed
to allow for an extensible evaluation benchmark. 



As illustrated in Figure~\ref{fig:summary_dashboard}, the calibration is done in two phases: (1)~at $t{=}0$, we \emph{train an initial IRT model} on a first dataset $B_0$ and a reference set of models, $\mathcal{M}_{ref}$, (2)~at each $t{>}0$ we integrate a new benchmark $B_t$ via \emph{sequential fixed parameter calibration}. To \emph{estimate model performance} at a given time point $t$, we use the corresponding set of anchors accumulated at that time.
Below, we elaborate on these different components.


\paragraph{Training an initial IRT model ($t=0$).}
We start by training an MIRT 2PL model on the first dataset~$B_0$, estimating parameters ($a$,~$d$; Equation~\ref{equation_1}) for all items in~$B_0$
via maximum likelihood over a set of responses from the reference models $\mathcal{M}_{\text{ref}}$. We then select a subset of anchor items $A_0 \subset B_0$ by clustering the IRT item representations and selecting the most representative item from each cluster, following the anchor selection method of~\citet{polo2024tinybenchmarksevaluatingllmsfewer}. These anchors will be used in all subsequent steps.

\paragraph{Sequential fixed parameter calibration ($t>0$).}
When a new benchmark~$B_t$ is added, we select anchors~$A_t$ while holding all existing anchor parameters fixed. To calculate the anchors~$A_t$ for this time step, we fit parameters for the items in~$B_t$ similarly to the initial IRT training. We use responses from $\mathcal{M}_{ref}$ on all items in~$B_t$, together with responses from the same models on the existing anchors~$A_{<t}$. Throughout this step, we constrain all existing anchor parameters to their previously estimated values (Equation~\ref{equation_1}). This yields a new set of anchor items~$A_t \subset B_t$, which is added to the accumulated anchor pool for all subsequent calibration steps. We denote the cumulative anchor set at step~$t$ as $A_{\leq t} = A_0 \cup A_1 \cup \cdots \cup A_t$.


\paragraph{Estimating model performance.}
To estimate a new model's performance at time step $t$, 
we estimate its ability
vector~$\boldsymbol{\theta}$ by maximizing the MIRT likelihood with
all item parameters fixed, 
given its responses to the set of anchors~$A_{\leq t}$:

\begin{equation}
\label{Equation_2}
\hat{\boldsymbol{\theta}} = \argmax_{\boldsymbol{\theta}}
\prod_{i \in A_{\leq t}} P(y_i \mid \boldsymbol{\theta};
\mathbf{a}_i, d_i)
\end{equation}

We then predict accuracy on each dataset $B_k$ using $\hat{\boldsymbol{\theta}}$ 
and the calibrated item parameters. Following the estimator 
of~\citet{polo2024tinybenchmarksevaluatingllmsfewer}, we combine the model's 
observed responses on the anchor items with the correctness probabilities 
that the IRT model assigns to all remaining items in $B_k$, weighting the 
two components to balance sampling variance against model bias. Evaluating a new model at step~$t$ requires responses to all accumulated anchors~$A_{\leq t}$, rather than to all items in the suite. For a previously evaluated model, each new dataset requires responses only to its newly selected anchors.


\section{Results: our approach enables efficient and extensible evaluation}

Here we extensively evaluate our framework on varying conditions (number of models and anchors) while simulating different orderings of dataset releases, in terms of estimation accuracy and computation cost.


\label{sec:results}
\subsection{Experimental setup}
\label{sec:experiments}

\begin{table*}[t]
    \centering
    \small
    \begin{tabular}{@{} l c c p{0.47\textwidth} @{}}
        \toprule
        \textbf{Benchmark Suite} & \textbf{\# Models} & \textbf{\# Datasets} & \textbf{Included Datasets (\# examples)} \\
        \midrule
        Open LLM Leaderboard & 395 & 6 &
          ARC Challenge (1{,}212), GSM8K (1{,}359), HellaSwag (10{,}082),
          MMLU (14{,}042), TruthfulQA (857), Winogrande (1{,}307) \\[3pt]
        MMLU & 428 & 57 &
          57 subdomains; (100--1{,}534 per subject) \\
        \bottomrule
    \end{tabular}
    \caption{Summary of the benchmark suites used in the experiments.}
    \label{tab:datasets}
\end{table*}

\begin{table}[tb!]
\centering
\small

\resizebox{0.47\columnwidth}{!}{%
\begin{tabular}{lrrr}
\toprule
\textbf{Dataset} & \textbf{Total Items} & \textbf{N=50} & \textbf{N=100} \\
\midrule
ARC Challenge  & 1{,}212  & 4.1\%  & 8.3\%  \\
GSM8K          & 1{,}359  & 3.7\%  & 7.4\%  \\
HellaSwag      & 10{,}082 & 0.5\%  & 1.0\%  \\
MMLU           & 14{,}042 & 0.4\%  & 0.7\%  \\
TruthfulQA     & 857      & 5.8\%  & 11.7\% \\
Winogrande     & 1{,}307  & 3.8\%  & 7.7\%  \\
\bottomrule
\end{tabular}%
}
\hfill
\resizebox{0.47\columnwidth}{!}{%
\begin{tabular}{lrrr}
\toprule
\textbf{Subject} & \textbf{Total Items} & \textbf{N=10} & \textbf{N=50} \\
\midrule
Abstract Algebra   & 100     & 10.0\% & 50.0\% \\
Computer Security  & 100     & 10.0\% & 50.0\% \\
Machine Learning   & 112     & 8.9\%  & 44.6\% \\
College Medicine   & 173     & 5.8\%  & 28.9\% \\
High School Math   & 270     & 3.7\%  & 18.5\% \\
Moral Scenarios    & 895     & 1.1\%  & 5.6\%  \\
Professional Law   & 1{,}534 & 0.7\%  & 3.3\%  \\
\midrule
\textit{Total} & \textit{14{,}042} & \textit{4.1\%} & \textit{20.3\%} \\
\bottomrule
\end{tabular}%
}

\caption{Anchor coverage as a percentage of total items for the Open LLM Leaderboard suite (left) and for representative subjects in the MMLU suite (right).}
\label{tab:anchor-coverage-both}
\end{table}



\label{sec:experiments}
We evaluate our framework on two benchmark suites, using randomized sequential ordering to simulate sequential benchmark addition over time.

\paragraph{Baselines.} We compare fixed parameter calibration against two baselines.
\textit{Concurrent calibration} jointly re-estimates all item parameters and model abilities each time a new benchmark is added. It uses the same IRT-based item selection procedure as fixed parameter calibration, but re-estimates all parameters on the accumulated data at each chain step without constraining any to prior values.
\textit{Random sampling} estimates performance by directly averaging accuracy on a randomly drawn subset of $N$ questions from the newly added dataset, without any IRT modeling.

\label{4.1_Datasets}
\paragraph{Datasets.}
Table~\ref{tab:datasets} summarizes the suites used to evaluate our approach. Both suites use item-level response data collected by \citet{polo2024tinybenchmarksevaluatingllmsfewer}. The first suite is the Open LLM Leaderboard~\citep{fourrier2024open}, covering six datasets. The second is the full MMLU~\citep{hendrycks2021measuringmassivemultitasklanguage}, treating its 57 subjects (e.g., algebra, law, medical genetics) as distinct datasets, which allows us to test stability over a longer sequence of benchmark additions to the evaluation suite.

\paragraph{Model splits.} We partition models into reference models, used for calibration, and held-out test models, in three ways, and apply all three to both suites. The random split assigns 75\% of models to the reference set and 25\% to the test set, following~\citet{polo2024tinybenchmarksevaluatingllmsfewer}. The time-ordered split orders models by their Open LLM Leaderboard submission date and assigns the earlier models to the reference set and the later ones to the test set. The family-held-out split assigns the Llama~\citep{touvron2023llamaopenefficientfoundation} and Mistral~\citep{jiang2023mistral7b} models to the test set and the rest to the reference set.


  
\paragraph{Dataset chain construction.} 

To simulate the gradual accumulation of datasets over time, we define a \emph{chain}: a sequence of datasets added one at a time to an initial suite. Each chain begins with a randomly selected subset of benchmarks forming the initial ``historic'' suite. At each subsequent \textit{chain step} $t$, one dataset is added and integrated via calibration; we then predict each test model's accuracy on the newly added dataset and compare it against its full-evaluation accuracy. Because we evaluate prediction quality at every step along the chain, each chain yields measurements at increasing chain lengths, allowing us to test whether prediction error accumulates as more benchmarks are added.

We sample multiple chains with different orderings per suite. For the Open LLM Leaderboard, each of the 6 benchmarks serves as the final (predicted) benchmark in turn, with 2 randomly sampled orderings of the preceding benchmarks per configuration (12 chains total). For MMLU, we sample 20 chains of sequentially added datasets, each ending with a different randomly selected subdomain from the 57 available. We report means and 95\% confidence intervals across chains throughout. Table~\ref{tab:anchor-coverage-both} contextualizes anchor sizes relative to benchmark scale, where $N$ denotes the number of anchor questions per benchmark.


\paragraph{Evaluation metrics.}

To assess the prediction quality, we use  mean absolute error (MAE):
\begin{equation}
    \text{MAE} = \frac{1}{|M_{\text{test}}|} \sum_{m \in M_{\text{test}}} |\widehat{\text{acc}}_m - \text{acc}_m|
\end{equation}
where $\text{acc}_m$ is the accuracy of model $m$ on the full benchmark and $\widehat{\text{acc}}_m$ is the estimated accuracy based on a subset of samples. Since leaderboard maintenance requires preserving relative model ordering, we also use Spearman's $\rho$ to measure ranking correlation.



\subsection{Results}
\label{sec:prediction_accuracy}
\begin{figure*}[tb!]
    \centering
    \includegraphics[width=\linewidth]{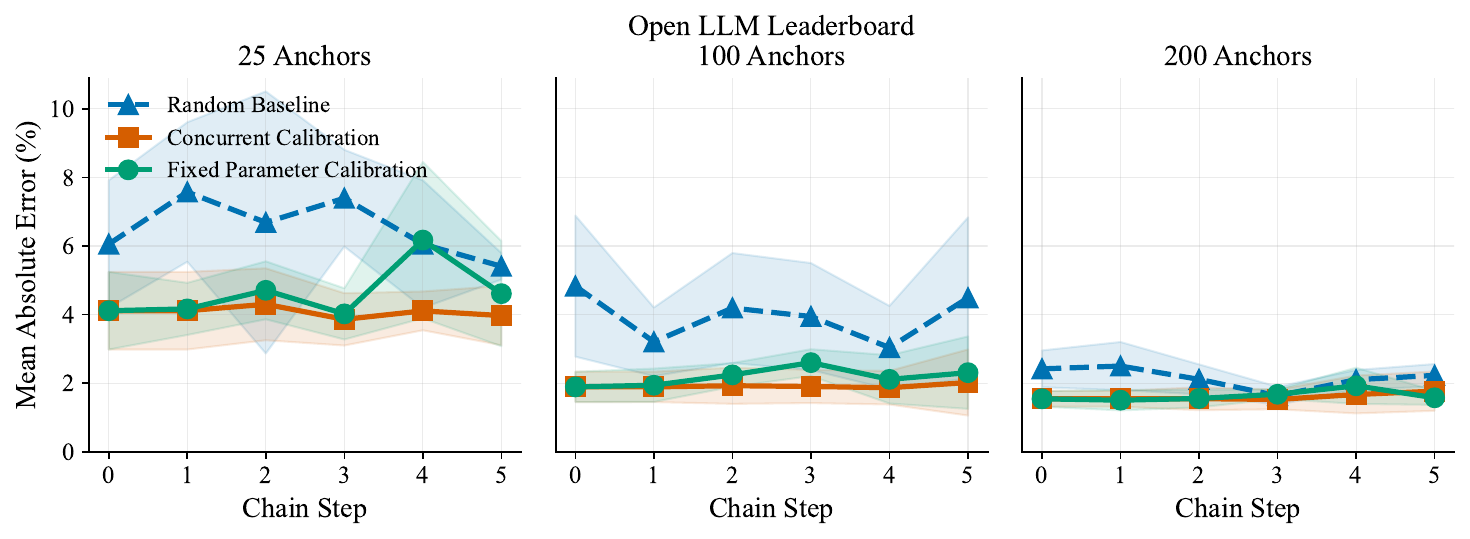}
    \vspace{0.8em}
    \includegraphics[width=\linewidth]{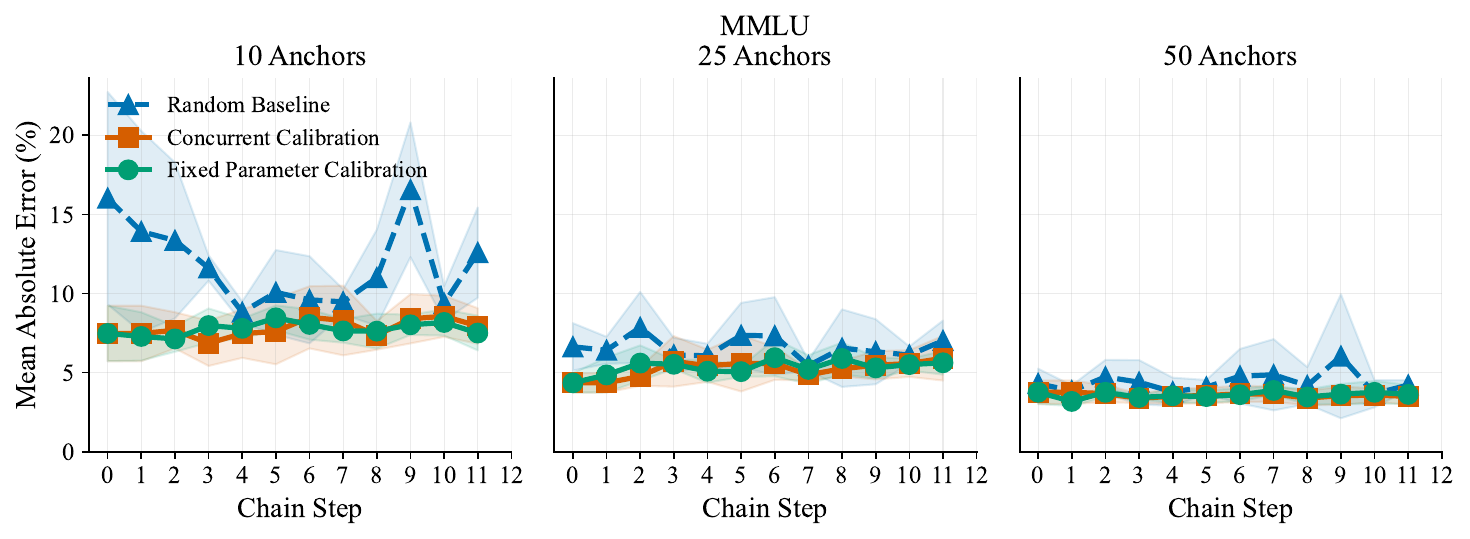}
    \caption{A small anchor budget suffices for accurate prediction across chain steps, with diminishing returns from larger sets. Across both benchmark suites, even compact sets (e.g., $N = 10$ or $25$) allow fixed parameter calibration and concurrent calibration to maintain low and stable MAE. As the number of anchors increases, random sampling approaches the performance of IRT-based methods, narrowing the gap between all three approaches. Shaded regions denote 95\% confidence intervals.}
    \label{fig:anchor_ablation}
\end{figure*}

\paragraph{Prediction error remains low as the benchmark suite grows.} Fixed parameter calibration provides strong predictive performance. As shown in Figure~\ref{fig:anchor_ablation}, its MAE remains low and stable across chain steps on both the Open LLM Leaderboard and MMLU, closely tracking concurrent calibration while consistently outperforming random sampling, especially at smaller anchor budgets. 

\begin{figure}[tb!]
    \centering
    \begin{minipage}[t]{0.5\columnwidth}
        \vspace{0pt}
        \includegraphics[width=\linewidth]{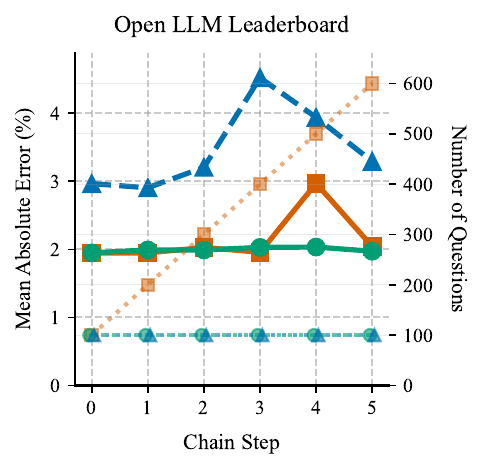}
    \end{minipage}
    \hfill
    \begin{minipage}[t]{0.49\columnwidth}
        \vspace{0pt}
        \includegraphics[width=\linewidth]{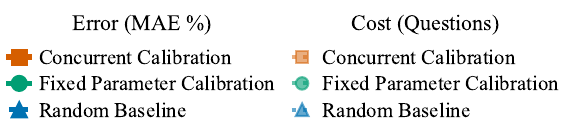}
\caption{
Fixed parameter calibration evaluates previously tested models only on the anchors of each newly added dataset, keeping the evaluation size constant as the suite grows. Concurrent calibration uses an increasingly large anchor set as the suite grows. Random sampling has the same fixed evaluation size but higher prediction error.}
        \label{fig:cost}
    \end{minipage}
\end{figure}

\begin{figure*}[tb!]
    \centering
    \includegraphics[width=\linewidth]{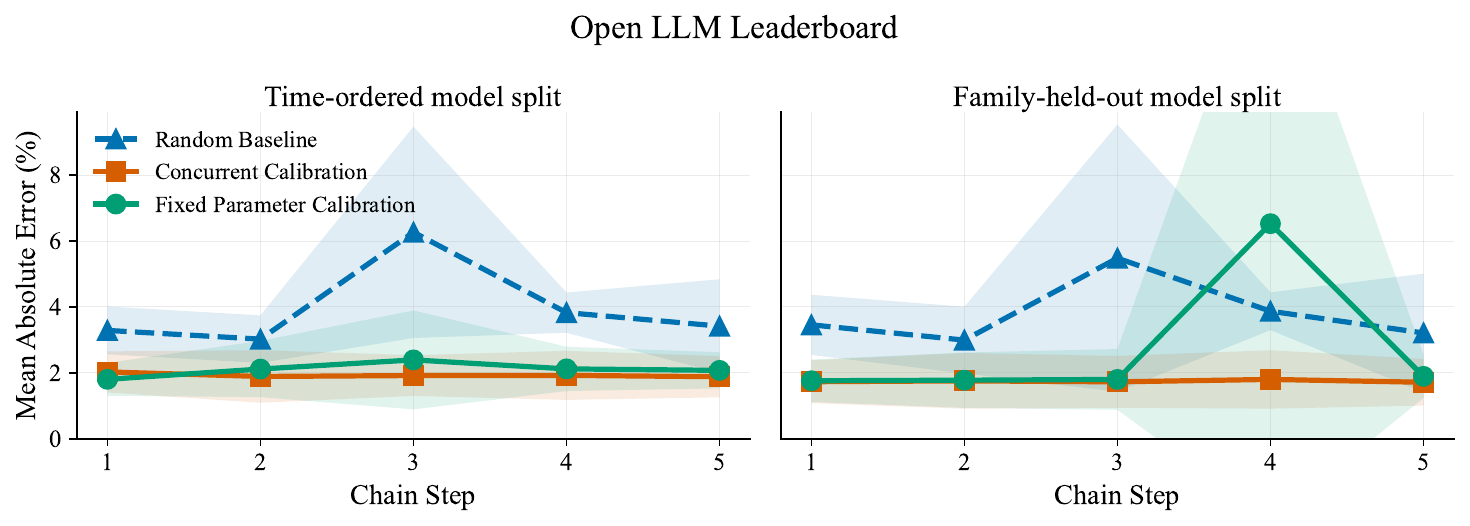}
    \caption{Prediction error remains low under time-ordered and family-held-out model splits on the Open LLM Leaderboard. Fixed parameter calibration tracks concurrent calibration and outperforms random sampling under both splits.}
    \label{fig:model_splits}
\end{figure*}

\paragraph{Prediction error remains low when models are split by time or family.}
Figure~\ref{fig:model_splits} shows low prediction error on the Open LLM Leaderboard under both splits. The time-ordered split uses earlier models as reference models and later models for testing, reflecting the arrival of new models over time. The family-held-out split removes the Llama and Mistral families from the reference set, testing performance on unseen model families. MMLU results show the same pattern and are reported in Appendix~\ref{app:realistic_model_splits}.

\begin{figure*}[tb!]
    \centering
    \includegraphics[width=\linewidth]{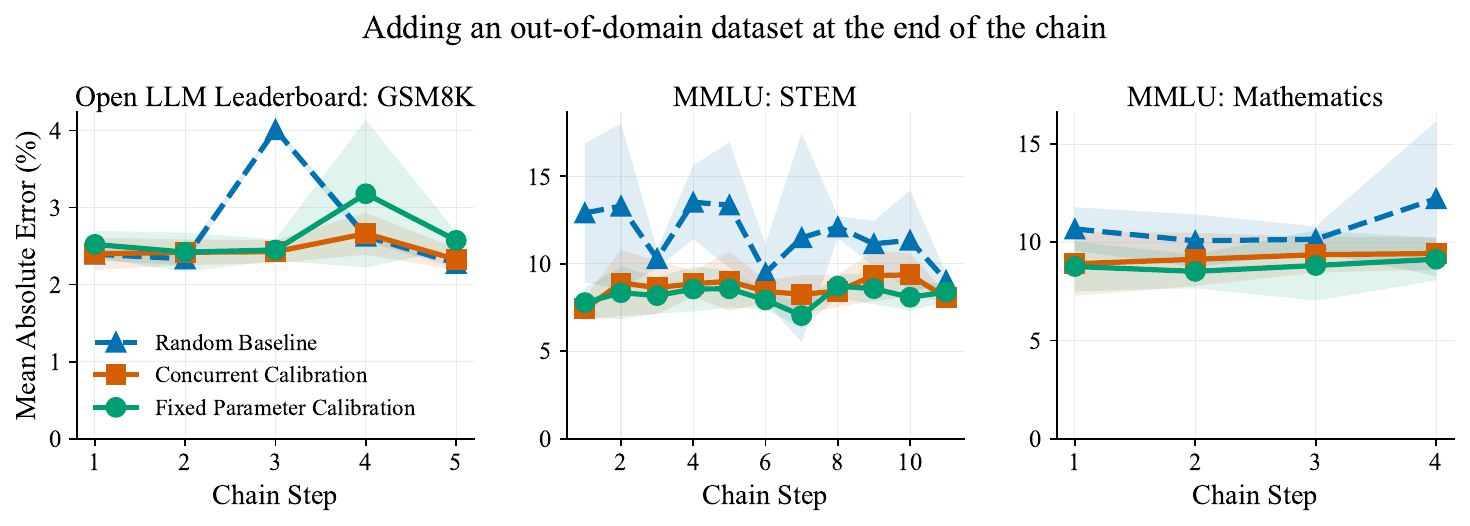}
    \caption{Prediction error stays low when an out-of-domain dataset is linked last, at increasing chain distance (the number of datasets added before it): GSM8K on the Open LLM Leaderboard (left), and the MMLU STEM (center) or mathematics (right) subjects. Fixed parameter calibration stays no worse than concurrent calibration and ahead of random sampling. Shaded regions denote 95\% confidence intervals.}
    \label{fig:ood_stress}
\end{figure*}

\paragraph{Prediction error remains stable when an out-of-domain dataset is added.}
To test performance when newly added datasets emphasize capabilities that are weakly represented earlier in the chain, we construct targeted orderings for the MMLU STEM and mathematics subjects and for GSM8K on the Open LLM Leaderboard. Figure~\ref{fig:ood_stress} shows that fixed parameter calibration matches or improves upon concurrent calibration on the MMLU settings and generally has lower prediction error than random sampling. On the Open LLM Leaderboard, its prediction error does not increase consistently as GSM8K is added later in the chain. These results show that fixed parameter calibration remains stable under the targeted orderings considered here.

\paragraph{Evaluation cost remains constant as the suite grows.} As shown in Figure~\ref{fig:cost}, when a new dataset is added, Fixed parameter calibration requires previously evaluated models to be run only on its anchors. Concurrent calibration reselects anchors across the accumulated suite at every step, requiring each model to be evaluated on the updated anchor set. As the suite grows, this anchor set becomes increasingly large.


\paragraph{The different approaches yield a good approximation of model rankings.}
Beyond absolute error, the approximation methods also recover the relative ordering of models well. As shown in Table~\ref{tab:ranking_both}, the Spearman correlations between predicted and full-evaluation rankings are strong. IRT-based approaches match or improve upon random sampling across configurations, with the clearest gains at low anchor counts. As the anchor budget grows, all methods move toward near-perfect ranking estimation.

\paragraph{Random selection is a viable approach for large enough anchor sets.}
Figure~\ref{fig:anchor_ablation} shows that although random sampling is  weaker than the calibrated methods in the low-budget regime, the gap narrows substantially as the anchor budget increases~\citep{perlitz2024efficientbenchmarkinglanguagemodels}.
However, the IRT-based methods retain a clear advantage at small anchor budgets, which is precisely the regime where evaluation cost savings are largest.

\begin{figure}[tb!]
    \centering
    \includegraphics[width=\linewidth]{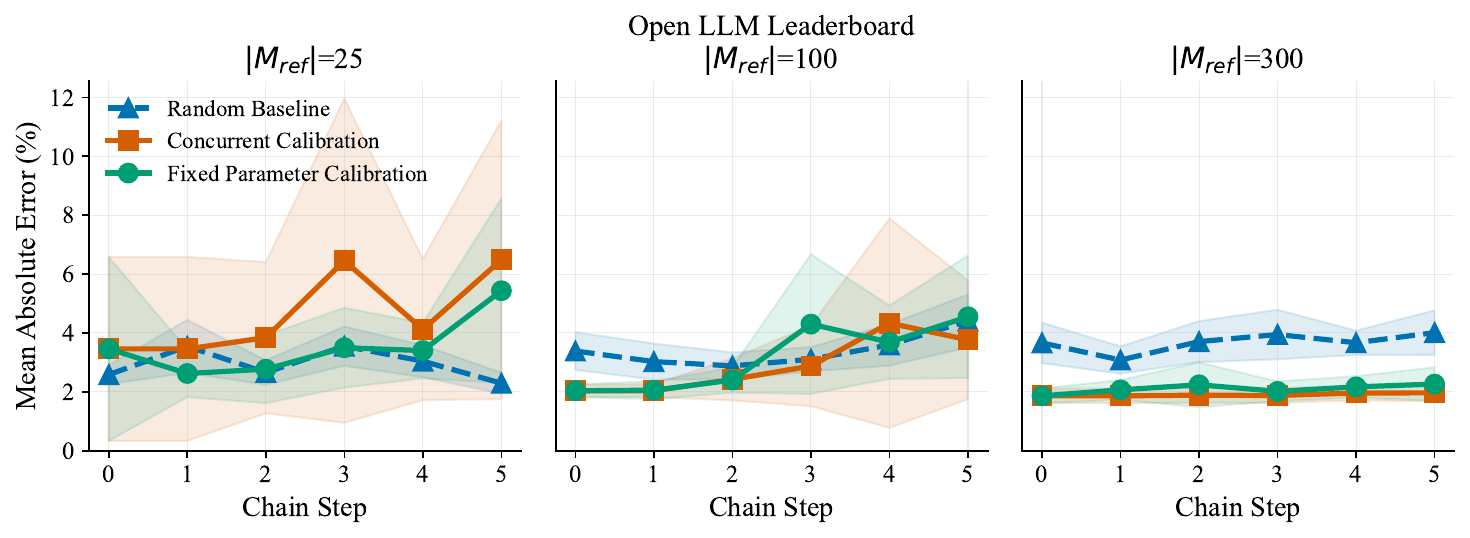}
    \vspace{0.8em}
    \includegraphics[width=\linewidth]{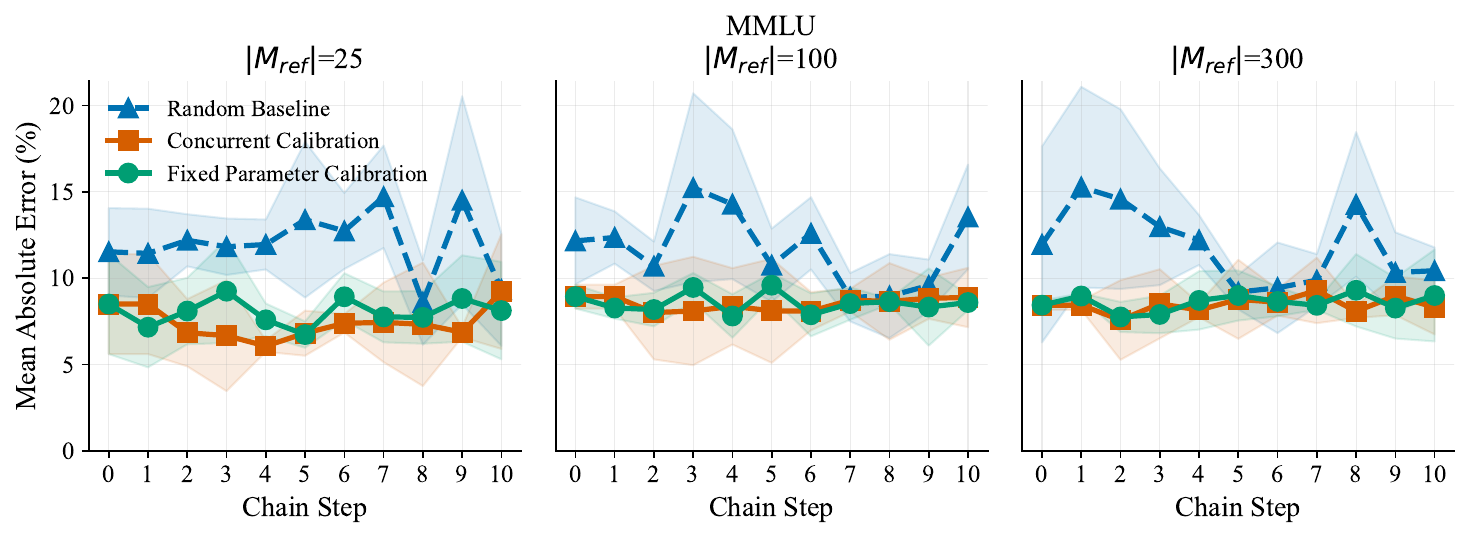}
    \caption{The effect of reference model count varies across suites. On the Open LLM Leaderboard (top), a reliable fixed parameter calibration is only achieved once the reference pool reaches 100 or more models. On MMLU (bottom), prediction quality is relatively robust even with only 25 reference models. Shaded regions denote 95\% confidence intervals.}
    \label{fig:model_sensitivity}
\end{figure}

\paragraph{Accurate approximation requires dozens of reference models, varying by suite.}
Figure~\ref{fig:model_sensitivity} shows that, under the random split, approximation quality depends on the number of reference models, and that a very small set of  models is insufficient for low prediction error. On the Open LLM Leaderboard, using only 25 reference models produces unstable error profiles for the IRT-based methods, while performance becomes reliable once the pool reaches roughly 100 models. On MMLU, the dependence is weaker, with 25 reference models already yielding robust performance, but the broader pattern is the same: accurate approximation requires at least dozens of reference models.

\newcolumntype{C}{>{\centering\arraybackslash}p{2.8em}}

\begin{table}[]
\centering
\begin{tabular}{@{} l *{3}{C} @{\hspace{1pt}} | @{\hspace{0.8pt}} *{4}{C} @{}}
\toprule
& \multicolumn{3}{c}{\textbf{Open LLM Leaderboard}} & \multicolumn{4}{c}{\textbf{MMLU}} \\
\cmidrule(lr){2-4} \cmidrule(lr){5-8}
\textit{Method} \textbackslash\ \textbf{Anchors ($N$)}
& \textbf{25} & \textbf{100} & \textbf{200} & \textbf{10} & \textbf{25} & \textbf{50} & \textbf{100} \\
\midrule
Random                                      & 0.82          & 0.91          & 0.96          & 0.68          & 0.85          & 0.91          & 0.98          \\
Concurrent Calibration                      & 0.89          & 0.94          & 0.97          & 0.73          & 0.83          & 0.90          & 0.98          \\
Fixed Parameter Calibration & 0.88 & 0.94 & 0.97 & 0.72 & 0.84 & 0.90 & 0.98 \\
\bottomrule
\end{tabular}
\caption{Ranking preservation (Spearman $\rho$) between predicted and full-evaluation model orderings. IRT-based calibration methods are competitive with random anchor sampling, with the clearest gains at small anchor counts.}
\label{tab:ranking_both}
\end{table}


\section{Related work}
\label{sec:related}
The escalating cost of LLM evaluation has driven research into efficient alternatives to comprehensive benchmarks~\citep{Luccioni_2024}. \cite{perlitz2024efficientbenchmarkinglanguagemodels} and~\cite{kipnis2025metabenchsparsebenchmark} showed that accurate rankings can be recovered from sparse item subsets. Predictive approaches go further, using IRT~\citep{polo2024tinybenchmarksevaluatingllmsfewer}, amortized modeling~\citep{truong2025reliableefficientamortizedmodelbased}, and multi-prompt estimators~\citep{polo2024efficientmultipromptevaluationllms} to infer full-benchmark performance from limited observations. While effective for static snapshots, these methods treat each benchmark as a closed system, lacking mechanisms for keeping results comparable as datasets evolve.

Psychometric theory offers a framework for such linkage.~\cite{lalor-etal-2016-building} introduced IRT to NLP for difficulty calibration,~\cite{rodriguez-etal-2021-evaluation} used it to build a Bayesian leaderboard model of ranking reliability within a single, closed benchmark, and~\cite{Li2025AdaptiveTF} apply adaptive testing to select informative questions dynamically. However, these efforts focus on ability \textit{estimation} within a fixed item pool, largely overlooking \textit{equating} across different test forms~\citep{Kolen2004TestES, https://doi.org/10.1002/j.2333-8504.1982.tb01311.x}.

Beyond equating, psychometric methods have also been applied to uncover the latent capability structure of LLMs. Factor-analytic studies have identified between three and eight interpretable skill dimensions that explain much of the variance in model performance across diverse tasks~\citep{burnell2023revealingstructurelanguagemodel, Ili__2024, maimon2026benchmarks}, and~\citet{polo2025slothscalinglawsllm} leverage similar latent-skill assumptions to model scaling laws across model families. These findings suggest that a low-dimensional skill space underlies LLM performance variation, consistent with the latent structure assumed by multidimensional IRT models.

To address benchmark saturation, researchers have proposed dynamic systems~\citep{kiela2021dynabenchrethinkingbenchmarkingnlp}, fluid evaluation~\citep{hofmann2025fluidlanguagemodelbenchmarking}, continuously refreshed benchmarks~\citep{livebench}, and aggregate-level unification of disparate metrics~\citep{ho2025rosettastoneaibenchmarks}. However, these approaches often fragment scores across evaluation periods; we address this by establishing comparability at the \textit{item level} using Multidimensional IRT, enabling extensible evaluation while preserving score comparability across evaluation periods.

\section{Discussion}
\label{sec:discussion}
A key assumption of our framework is that the latent dimensions identified during initial calibration generalize to subsequently added benchmarks. Our results support this: the flat MAE profiles across both suites show that the prediction
accuracy does not degrade as the benchmark pool expands. This stability follows from the design of fixed parameter calibration, which holds anchor parameters fixed while estimating new item parameters ($\mathbf{a}_i$, $d_i$) freely at each chain step. This allows new benchmarks to be integrated while preserving the existing scale. In practice, this framework could be applied to leaderboards such as the Open LLM Leaderboard, allowing new datasets and models to be integrated efficiently as the benchmark suite evolves.

However, prediction quality depends on the degree of overlap between new benchmarks and the existing latent space. When a new benchmark tests a capability largely absent from existing tasks, we expect prediction accuracy to degrade. This is not specific to our approach; any equating procedure requires shared latent structure across old and new items. In practice, if a new capability proves important to the evaluation community, additional benchmarks probing it are likely to follow. As these benchmarks are calibrated into the chain, the latent space gradually gains coverage of that skill, and prediction quality recovers. We hypothesize that MMLU's 57 subjects share more latent structure, as they are all multiple-choice knowledge questions, making calibration effective with fewer reference models. The Open LLM Leaderboard's six datasets span more diverse task types, requiring a larger reference pool to capture the relevant latent dimensions.

This motivates a natural extension: organizing benchmarks into skill clusters, each maintaining its own calibrated parameters. New benchmarks could be routed to an existing cluster or, when prediction quality drops substantially, initiate a new calibration chain for a previously unrepresented capability. Extending the framework to support skill-based clustering is a promising direction for future work, raising questions including how to define clusters from item-level data, how to route incoming benchmarks to existing or new clusters, and how to model interactions between separately calibrated parameter sets.

\section{Conclusion}
\label{sec:conclusion}

Evaluating LLMs consistently has become harder as models and benchmarks are released continuously. We address this problem by casting evaluation under evolving benchmark coverage as a scale-linking problem and introducing a multidimensional IRT framework with anchor items and fixed parameter calibration. Experiments on the Open LLM Leaderboard and MMLU achieve prediction error of 2–3\% using only 100 anchor questions per dataset, with no error accumulation across long calibration chains. The framework supports efficient evaluation of new models and integration of newly added datasets while preserving score comparability.

\section{Limitations}
\label{sec:limitations}
First, our empirical validation is limited to knowledge and reasoning tasks in English on the Open LLM Leaderboard and MMLU, and the framework's behavior on other task types or languages remains to be tested. Second, the current framework operates on binary item responses, and extending it to graded or open-ended evaluation formats would require adaptations to the IRT formulation. Third, although we evaluate time-ordered and family-held-out splits in addition to random partitioning, future models may behave differently. Our targeted dataset orderings also use datasets for which response data is already available across many models, and other datasets may behave differently when added to the chain. Finally, fixed parameter calibration assumes that anchor questions maintain stable statistical properties as models evolve. Anchors may become contaminated or saturated over time, requiring recalibration. Evaluating a new model also requires responses to all accumulated anchors, so its evaluation cost grows as the suite expands, although it remains substantially lower than full-suite inference.

\section*{Acknowledgments}
This research was conducted in collaboration with the Hebrew University of Jerusalem and IBM Research. The work was supported by the IBM-HUJI Research collaboration. This work was partially supported by research grant
no. 7256 from the Israeli Ministry of Science and
Technology. This work was also partially supported by the PBC Scholarship for Outstanding Doctoral Students from Diverse Populations and the Socioeconomic Periphery.

\bibliography{custom}

@article{maimon2026benchmarks,
  title={From Benchmarks to Skills: Low-Rank Factors for LLM Evaluation},
  author={Maimon, Aviya and Cohen, Amir David Nisan and Vishne, Gal and Ravfogel, Shauli and Tsarfaty, Reut},
  journal={Transactions of the Association for Computational Linguistics},
  volume={14},
  pages={1639--1653},
  year={2026},
  publisher={MIT Press 255 Main Street, 9th Floor, Cambridge, Massachusetts 02142, USA~…}
}

@misc{polo2024tinybenchmarksevaluatingllmsfewer,
      title={tinyBenchmarks: evaluating LLMs with fewer examples}, 
      author={Felipe Maia Polo and Lucas Weber and Leshem Choshen and Yuekai Sun and Gongjun Xu and Mikhail Yurochkin},
      year={2024},
      eprint={2402.14992},
      archivePrefix={arXiv},
      primaryClass={cs.CL},
      url={https://arxiv.org/abs/2402.14992}, 
}

@misc{ho2025rosettastoneaibenchmarks,
 author = {Anson Ho and Jean-Stanislas Denain and David Atanasov and Samuel Albanie and Rohin Shah},
 journal = {ArXiv preprint},
 title = {A Rosetta Stone for AI Benchmarks},
 url = {https://arxiv.org/abs/2512.00193},
 volume = {abs/2512.00193},
 year = {2025}
}

@misc{hofmann2025fluidlanguagemodelbenchmarking,
 author = {Valentin Hofmann and David Heineman and Ian Magnusson and Kyle Lo and Jesse Dodge and Maarten Sap and Pang Wei Koh and Chun Wang and Hannaneh Hajishirzi and Noah A. Smith},
 journal = {ArXiv preprint},
 title = {Fluid Language Model Benchmarking},
 url = {https://arxiv.org/abs/2509.11106},
 volume = {abs/2509.11106},
 year = {2025}
}

@inproceedings{kiela2021dynabenchrethinkingbenchmarkingnlp,
 address = {Online},
 author = {Kiela, Douwe  and
Bartolo, Max  and
Nie, Yixin  and
Kaushik, Divyansh  and
Geiger, Atticus  and
Wu, Zhengxuan  and
Vidgen, Bertie  and
Prasad, Grusha  and
Singh, Amanpreet  and
Ringshia, Pratik  and
Ma, Zhiyi  and
Thrush, Tristan  and
Riedel, Sebastian  and
Waseem, Zeerak  and
Stenetorp, Pontus  and
Jia, Robin  and
Bansal, Mohit  and
Potts, Christopher  and
Williams, Adina},
 booktitle = {Proceedings of the 2021 Conference of the North American Chapter of the Association for Computational Linguistics: Human Language Technologies},
 doi = {10.18653/v1/2021.naacl-main.324},
 editor = {Toutanova, Kristina  and
Rumshisky, Anna  and
Zettlemoyer, Luke  and
Hakkani-Tur, Dilek  and
Beltagy, Iz  and
Bethard, Steven  and
Cotterell, Ryan  and
Chakraborty, Tanmoy  and
Zhou, Yichao},
 pages = {4110--4124},
 publisher = {Association for Computational Linguistics},
 title = {Dynabench: Rethinking Benchmarking in {NLP}},
 url = {https://aclanthology.org/2021.naacl-main.324},
 year = {2021}
}

@inproceedings{Kolen2004TestES,
 author = {Michael J. Kolen and Robert L. Brennan},
 title = {Test Equating, Scaling, and Linking: Methods and Practices},
 url = {https://api.semanticscholar.org/CorpusID:119066024},
 year = {2004}
}

@article{Li2025AdaptiveTF,
 author = {Peiyu Li and Xiuxiu Tang and Si Chen and Ying Cheng and Ronald A. Metoyer and Ting Hua and Nitesh V. Chawla},
 journal = {ArXiv preprint},
 title = {Adaptive Testing for LLM Evaluation: A Psychometric Alternative to Static Benchmarks},
 url = {https://arxiv.org/abs/2511.04689},
 volume = {abs/2511.04689},
 year = {2025}
}

@inproceedings{polo2024efficientmultipromptevaluationllms,
 author = {Felipe Maia Polo and
Ronald Xu and
Lucas Weber and
M{\'{\i}}rian Silva and
Onkar Bhardwaj and
Leshem Choshen and
Allysson Flavio Melo de Oliveira and
Yuekai Sun and
Mikhail Yurochkin},
 bibsource = {dblp computer science bibliography, https://dblp.org},
 booktitle = {Advances in Neural Information Processing Systems 38: Annual Conference
on Neural Information Processing Systems 2024, NeurIPS 2024, Vancouver,
BC, Canada, December 10 - 15, 2024},
 editor = {Amir Globersons and
Lester Mackey and
Danielle Belgrave and
Angela Fan and
Ulrich Paquet and
Jakub M. Tomczak and
Cheng Zhang},
 title = {Efficient multi-prompt evaluation of LLMs},
 url = {http://papers.nips.cc/paper\_files/paper/2024/hash/28236482f64a72eec43706b6f3a6c511-Abstract-Conference.html},
 year = {2024}
}

@misc{truong2025reliableefficientamortizedmodelbased,
 author = {Sang Truong and Yuheng Tu and Percy Liang and Bo Li and Sanmi Koyejo},
 journal = {ArXiv preprint},
 title = {Reliable and Efficient Amortized Model-based Evaluation},
 url = {https://arxiv.org/abs/2503.13335},
 volume = {abs/2503.13335},
 year = {2025}
}

@misc{kipnis2025metabenchsparsebenchmark,
 author = {Alex Kipnis and Konstantinos Voudouris and Luca M. Schulze Buschoff and Eric Schulz},
 journal = {ArXiv preprint},
 title = {metabench -- A Sparse Benchmark of Reasoning and Knowledge in Large Language Models},
 url = {https://arxiv.org/abs/2407.12844},
 volume = {abs/2407.12844},
 year = {2024}
}

@article{article,
 author = {Baker, F. and Kim, S.},
 pages = {},
 title = {Item Response Theory: Parameter Estimation Techniques},
 year = {2004}
}

@article{kim1998comparison,
  title={A comparison of linking and concurrent calibration under item response theory},
  author={Kim, Seock-Ho and Cohen, Allan S},
  journal={Applied psychological measurement},
  volume={22},
  number={2},
  pages={131--143},
  year={1998},
  publisher={SAGE PUBLICATIONS, INC. 2455 Teller Road, Thousand Oaks, CA 91320}
}

@article{Lord1984ComparisonOI,
 author = {Frederic M. Lord and Marilyn S. Wingersky},
 journal = {Applied Psychological Measurement},
 pages = {453 - 461},
 title = {Comparison of IRT True-Score and Equipercentile Observed-Score "Equatings"},
 url = {https://api.semanticscholar.org/CorpusID:121685628},
 volume = {8},
 year = {1984}
}

@misc{hendrycks2021measuringmassivemultitasklanguage,
      title={Measuring Massive Multitask Language Understanding}, 
      author={Dan Hendrycks and Collin Burns and Steven Basart and Andy Zou and Mantas Mazeika and Dawn Song and Jacob Steinhardt},
      year={2021},
      eprint={2009.03300},
      archivePrefix={arXiv},
      primaryClass={cs.CY},
      url={https://arxiv.org/abs/2009.03300}, 
}

@misc{perlitz2024llmbenchmarksagreefixing,
 author = {Yotam Perlitz and Ariel Gera and Ofir Arviv and Asaf Yehudai and Elron Bandel and Eyal Shnarch and Michal Shmueli-Scheuer and Leshem Choshen},
 journal = {ArXiv preprint},
 title = {Do These LLM Benchmarks Agree? Fixing Benchmark Evaluation with BenchBench},
 url = {https://arxiv.org/abs/2407.13696},
 volume = {abs/2407.13696},
 year = {2024}
}

@inproceedings{perlitz2024efficientbenchmarkinglanguagemodels,
 address = {Mexico City, Mexico},
 author = {Perlitz, Yotam  and
Bandel, Elron  and
Gera, Ariel  and
Arviv, Ofir  and
Ein-Dor, Liat  and
Shnarch, Eyal  and
Slonim, Noam  and
Shmueli-Scheuer, Michal  and
Choshen, Leshem},
 booktitle = {Proceedings of the 2024 Conference of the North American Chapter of the Association for Computational Linguistics: Human Language Technologies (Volume 1: Long Papers)},
 editor = {Duh, Kevin  and
Gomez, Helena  and
Bethard, Steven},
 pages = {2519--2536},
 publisher = {Association for Computational Linguistics},
 title = {Efficient Benchmarking (of Language Models)},
 url = {https://aclanthology.org/2024.naacl-long.139},
 year = {2024}
}

@article{https://doi.org/10.1002/j.2333-8504.1982.tb01311.x,
 author = {Stocking, Martha L. and Lord, Frederic M.},
 doi = {https://doi.org/10.1002/j.2333-8504.1982.tb01311.x},
 eprint = {https://onlinelibrary.wiley.com/doi/pdf/10.1002/j.2333-8504.1982.tb01311.x},
 journal = {ETS Research Report Series},
 number = {1},
 pages = {i-29},
 title = {DEVELOPING A COMMON METRIC IN ITEM RESPONSE THEORY},
 url = {https://onlinelibrary.wiley.com/doi/abs/10.1002/j.2333-8504.1982.tb01311.x},
 volume = {1982},
 year = {1982}
}

@inproceedings{livebench,
 author = {Colin White and Samuel Dooley and Manley Roberts and Arka Pal and Benjamin Feuer and Siddhartha Jain and Ravid Shwartz-Ziv and Neel Jain and Khalid Saifullah and Sreemanti Dey and Shubh-Agrawal and Sandeep Singh Sandha and Siddartha Venkat Naidu and Chinmay Hegde and Yann LeCun and Tom Goldstein and Willie Neiswanger and Micah Goldblum},
 booktitle = {The Thirteenth International Conference on Learning Representations},
 title = {LiveBench: A Challenging, Contamination-Free {LLM} Benchmark},
 year = {2025}
}

@inproceedings{Luccioni_2024,
 author = {Luccioni, Sasha and Jernite, Yacine and Strubell, Emma},
 booktitle = {The 2024 ACM Conference on Fairness Accountability and Transparency},
 collection = {FAccT ’24},
 doi = {10.1145/3630106.3658542},
 publisher = {ACM},
 series = {FAccT ’24},
 title = {Power Hungry Processing: Watts Driving the Cost of AI Deployment?},
 url = {http://dx.doi.org/10.1145/3630106.3658542},
 year = {2024}
}

@inproceedings{rodriguez-etal-2021-evaluation,
 address = {Online},
 author = {Rodriguez, Pedro  and
Barrow, Joe  and
Hoyle, Alexander Miserlis  and
Lalor, John P.  and
Jia, Robin  and
Boyd-Graber, Jordan},
 booktitle = {Proceedings of the 59th Annual Meeting of the Association for Computational Linguistics and the 11th International Joint Conference on Natural Language Processing (Volume 1: Long Papers)},
 doi = {10.18653/v1/2021.acl-long.346},
 editor = {Zong, Chengqing  and
Xia, Fei  and
Li, Wenjie  and
Navigli, Roberto},
 pages = {4486--4503},
 publisher = {Association for Computational Linguistics},
 title = {Evaluation Examples are not Equally Informative: How should that change {NLP} Leaderboards?},
 url = {https://aclanthology.org/2021.acl-long.346},
 year = {2021}
}

@misc{polo2025slothscalinglawsllm,
 author = {Felipe Maia Polo and Seamus Somerstep and Leshem Choshen and Yuekai Sun and Mikhail Yurochkin},
 journal = {ArXiv preprint},
 title = {Sloth: scaling laws for LLM skills to predict multi-benchmark performance across families},
 url = {https://arxiv.org/abs/2412.06540},
 volume = {abs/2412.06540},
 year = {2024}
}

@misc{burnell2023revealingstructurelanguagemodel,
      title={Revealing the structure of language model capabilities}, 
      author={Ryan Burnell and Han Hao and Andrew R. A. Conway and Jose Hernandez Orallo},
      year={2023},
      eprint={2306.10062},
      archivePrefix={arXiv},
      primaryClass={cs.CL},
      url={https://arxiv.org/abs/2306.10062}, 
}

@inproceedings{lalor-etal-2016-building,
    title = "Building an Evaluation Scale using Item Response Theory",
    author = "Lalor, John P.  and
      Wu, Hao  and
      Yu, Hong",
    editor = "Su, Jian  and
      Duh, Kevin  and
      Carreras, Xavier",
    booktitle = "Proceedings of the 2016 Conference on Empirical Methods in Natural Language Processing",
    month = nov,
    year = "2016",
    address = "Austin, Texas",
    publisher = "Association for Computational Linguistics",
    url = "https://aclanthology.org/D16-1062/",
    doi = "10.18653/v1/D16-1062",
    pages = "648--657"
}

@misc{fourrier2024open,
  title={Open llm leaderboard v2},
  author={Fourrier, Cl{\'e}mentine and Habib, Nathan and Lozovskaya, Alina and Szafer, Konrad and Wolf, Thomas},
  year={2024}
}

@misc{touvron2023llamaopenefficientfoundation,
      title={LLaMA: Open and Efficient Foundation Language Models}, 
      author={Hugo Touvron and Thibaut Lavril and Gautier Izacard and Xavier Martinet and Marie-Anne Lachaux and Timothée Lacroix and Baptiste Rozière and Naman Goyal and Eric Hambro and Faisal Azhar and Aurelien Rodriguez and Armand Joulin and Edouard Grave and Guillaume Lample},
      year={2023},
      eprint={2302.13971},
      archivePrefix={arXiv},
      primaryClass={cs.CL},
      url={https://arxiv.org/abs/2302.13971}, 
}

@misc{jiang2023mistral7b,
      title={Mistral 7B}, 
      author={Albert Q. Jiang and Alexandre Sablayrolles and Arthur Mensch and Chris Bamford and Devendra Singh Chaplot and Diego de las Casas and Florian Bressand and Gianna Lengyel and Guillaume Lample and Lucile Saulnier and Lélio Renard Lavaud and Marie-Anne Lachaux and Pierre Stock and Teven Le Scao and Thibaut Lavril and Thomas Wang and Timothée Lacroix and William El Sayed},
      year={2023},
      eprint={2310.06825},
      archivePrefix={arXiv},
      primaryClass={cs.CL},
      url={https://arxiv.org/abs/2310.06825}, 
}

@article{Ili__2024,
   title={Evidence of interrelated cognitive-like capabilities in large language models: Indications of artificial general intelligence or achievement?},
   volume={106},
   ISSN={0160-2896},
   url={http://dx.doi.org/10.1016/j.intell.2024.101858},
   DOI={10.1016/j.intell.2024.101858},
   journal={Intelligence},
   publisher={Elsevier BV},
   author={Ilić, David and Gignac, Gilles E.},
   year={2024},
   month=Sept, pages={101858} }
\bibliographystyle{colm2026_conference}

\clearpage
\appendix

\section{Anchor selection analysis}
\label{app:item_maps}
\begin{figure}[h]
    \centering
    \includegraphics[width=\columnwidth]{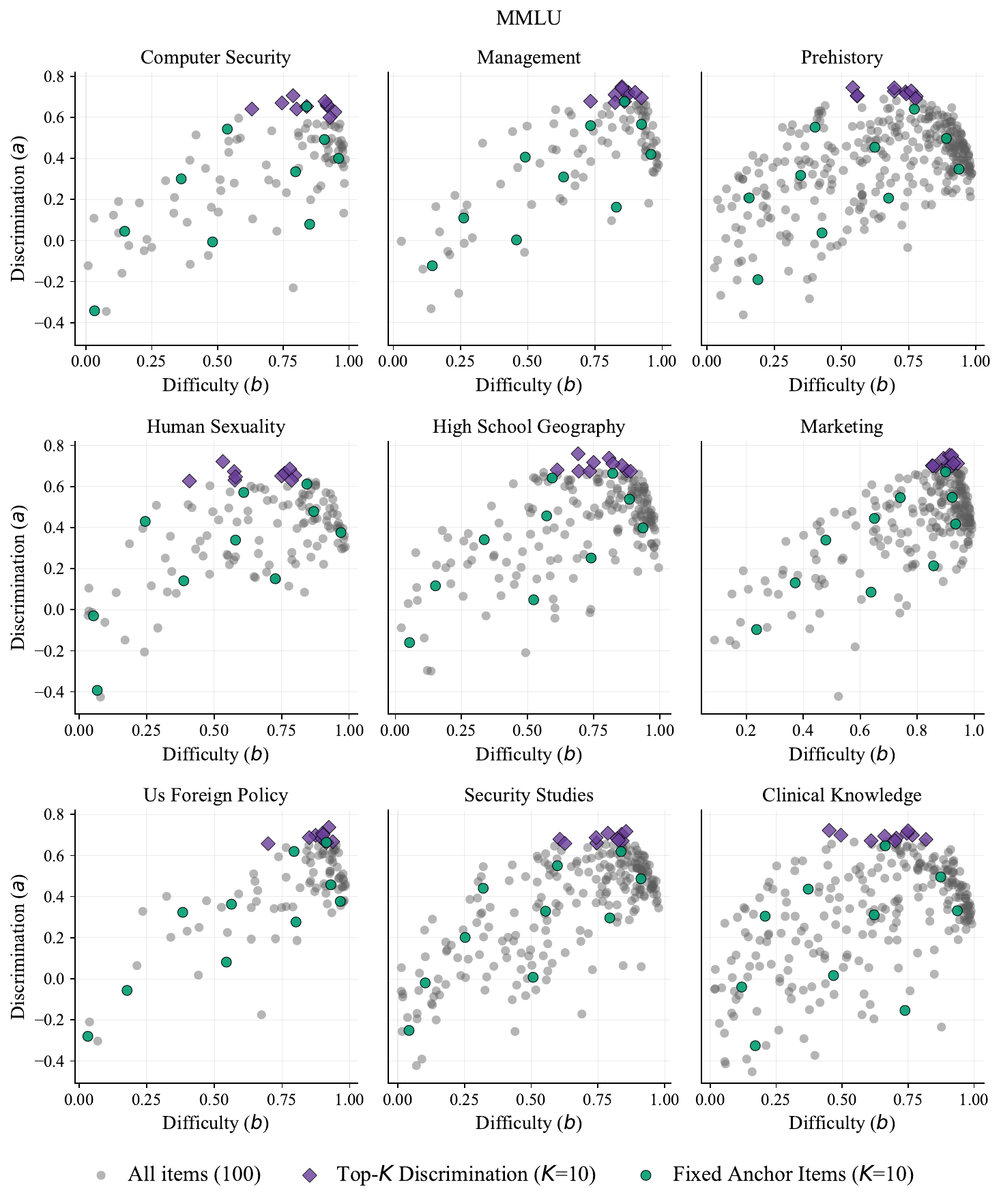}
    \caption{Item maps showing the difficulty ($b$) and discrimination ($a$) of all items,
    with selected anchor items highlighted. Clustering-based selection (circles) distributes
    anchors across the full parameter space, while top-$K$ selection (diamonds) concentrates
    them in a narrow high-discrimination region. Representative MMLU subjects.}
    \label{fig:item_map_overlay_mmlu}
\end{figure}

\begin{figure}[t!]
    \centering
    \includegraphics[width=\columnwidth]{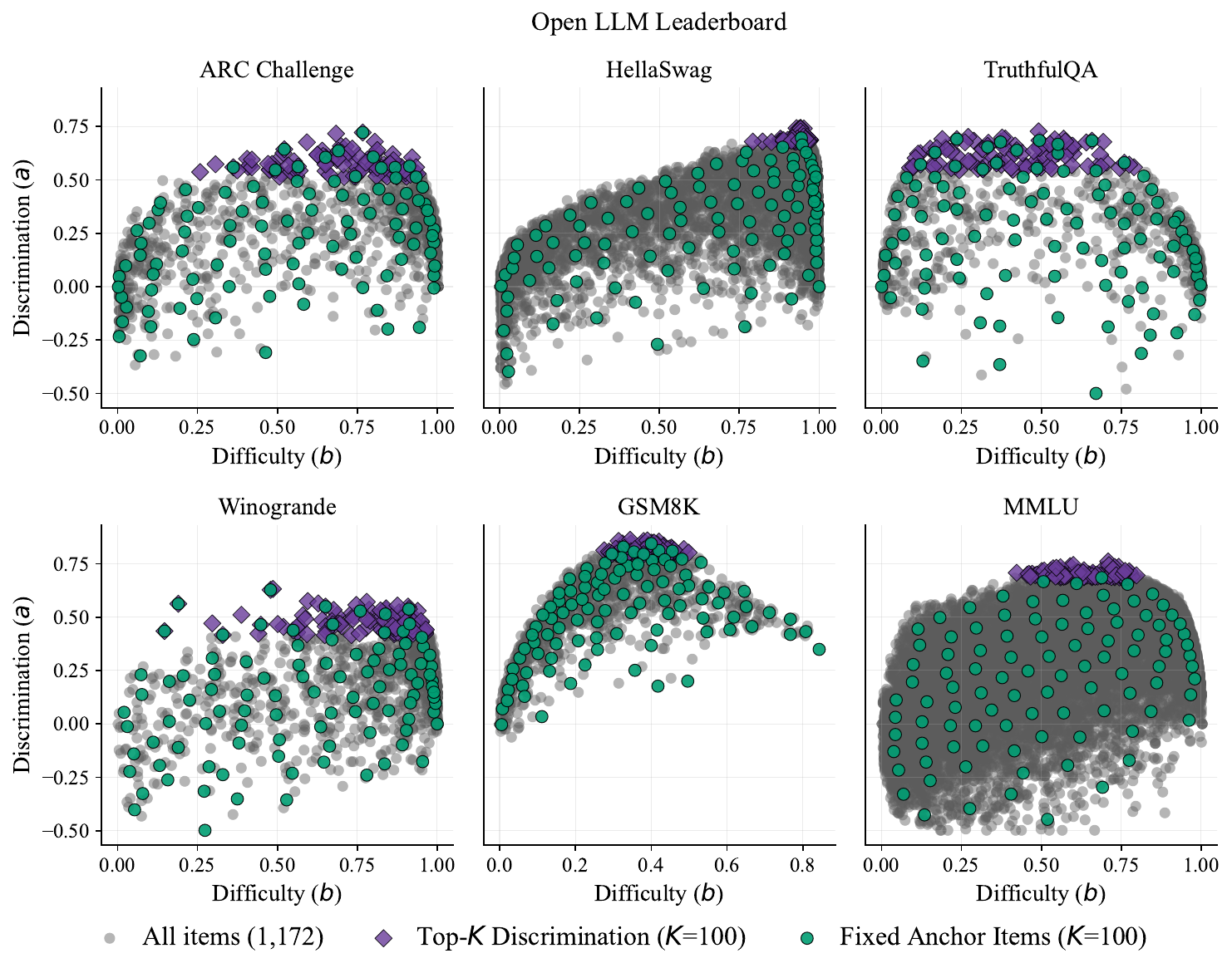}
    \caption{Item maps showing the difficulty ($b$) and discrimination ($a$) of all items,
    with selected anchor items highlighted. Clustering-based selection (circles) distributes
    anchors across the full parameter space, while top-$K$ selection (diamonds) concentrates
    them in a narrow high-discrimination region. Open LLM Leaderboard datasets.}
    \label{fig:item_map_overlay_lb}
\end{figure}

\begin{figure}[tb!]
    \centering
    \begin{minipage}[t]{0.5\columnwidth}
        \vspace{0pt}
        \includegraphics[width=\linewidth]{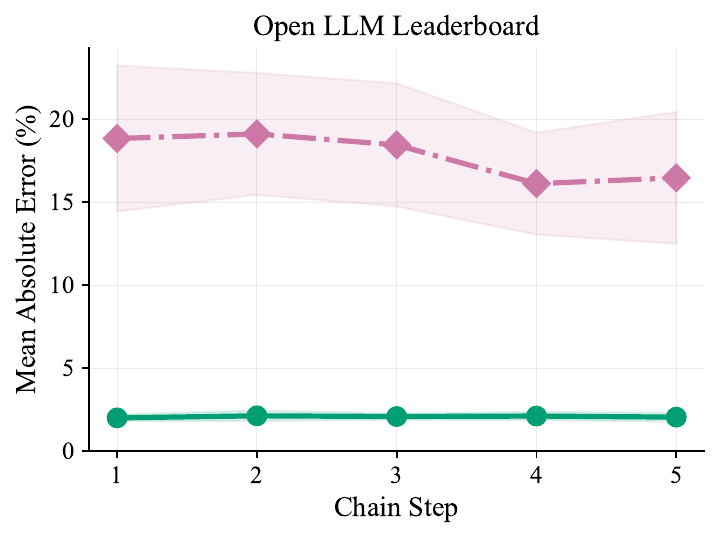}
    \end{minipage}
    \hfill
    \begin{minipage}[t]{0.49\columnwidth}
        \vspace{0pt}
        \includegraphics[width=\linewidth]{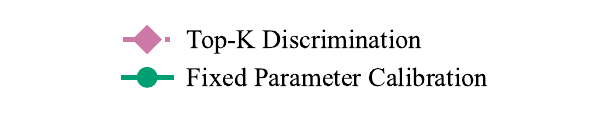}
        
        \caption{Replacing the clustering-based anchor selection with top-$K$ selection by discrimination parameter leads to substantially higher MAE on the Open LLM Leaderboard, while the rest of the fixed parameter calibration pipeline remains identical. Shaded regions denote 95\% confidence intervals.}
        \label{fig:topk_ablation}
    \end{minipage}
\end{figure}

\begin{figure}[tb!]
    \centering
    \begin{minipage}[t]{0.5\columnwidth}
        \vspace{0pt}
        \includegraphics[width=\linewidth]{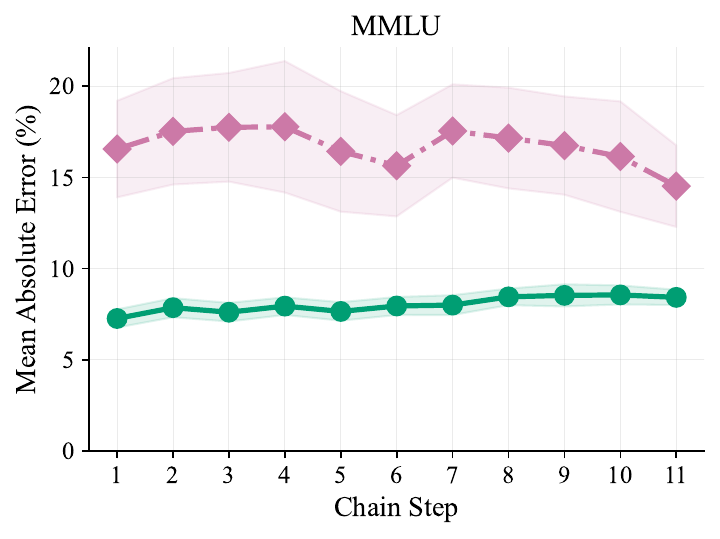}
    \end{minipage}
    \hfill
    \begin{minipage}[t]{0.49\columnwidth}
        \vspace{0pt}
        \includegraphics[width=\linewidth]{figures_next_arr/fig_topk_ablation_legend_vertical.pdf}
        
        \caption{Replacing the clustering-based anchor selection with top-$K$ selection by discrimination parameter leads to substantially higher MAE on MMLU, while the rest of the fixed parameter calibration pipeline remains identical. Shaded regions denote 95\% confidence intervals.}
        \label{fig:topk_ablation_mmlu}
    \end{minipage}
\end{figure}

Figures~\ref{fig:item_map_overlay_mmlu} and~\ref{fig:item_map_overlay_lb} visualize the item representations and the anchors selected by our clustering-based procedure on representative MMLU subjects and the Open LLM Leaderboard datasets. We additionally compare this procedure with a top-K baseline that selects the K items with the highest discrimination. As shown in Figures~\ref{fig:topk_ablation} and~\ref{fig:topk_ablation_mmlu}, top-K selection produces substantially higher prediction error on both benchmark suites. The item maps show that clustering distributes anchors across the parameter space, while top-K selection concentrates them in a narrow high-discrimination region.

\section{Additional results under realistic model splits}
\label{app:realistic_model_splits}
\begin{figure*}[tb!]
    \centering
    \includegraphics[width=\linewidth]{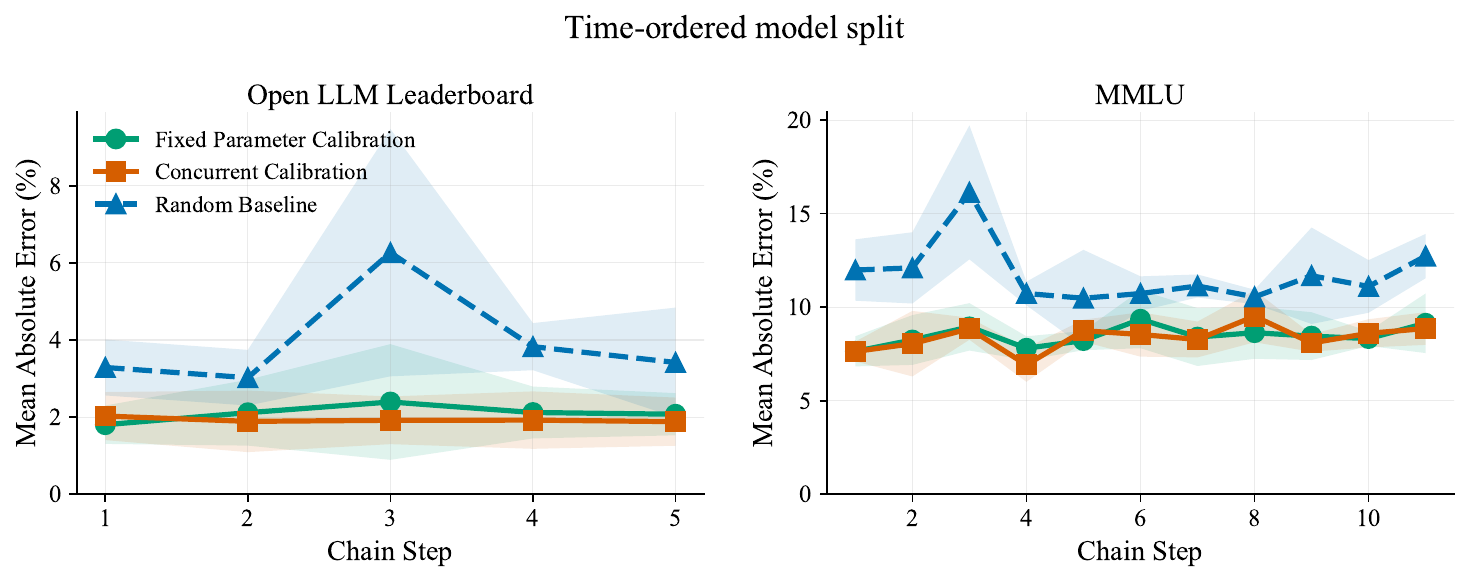}
    \vspace{0.8em}
    \includegraphics[width=\linewidth]    {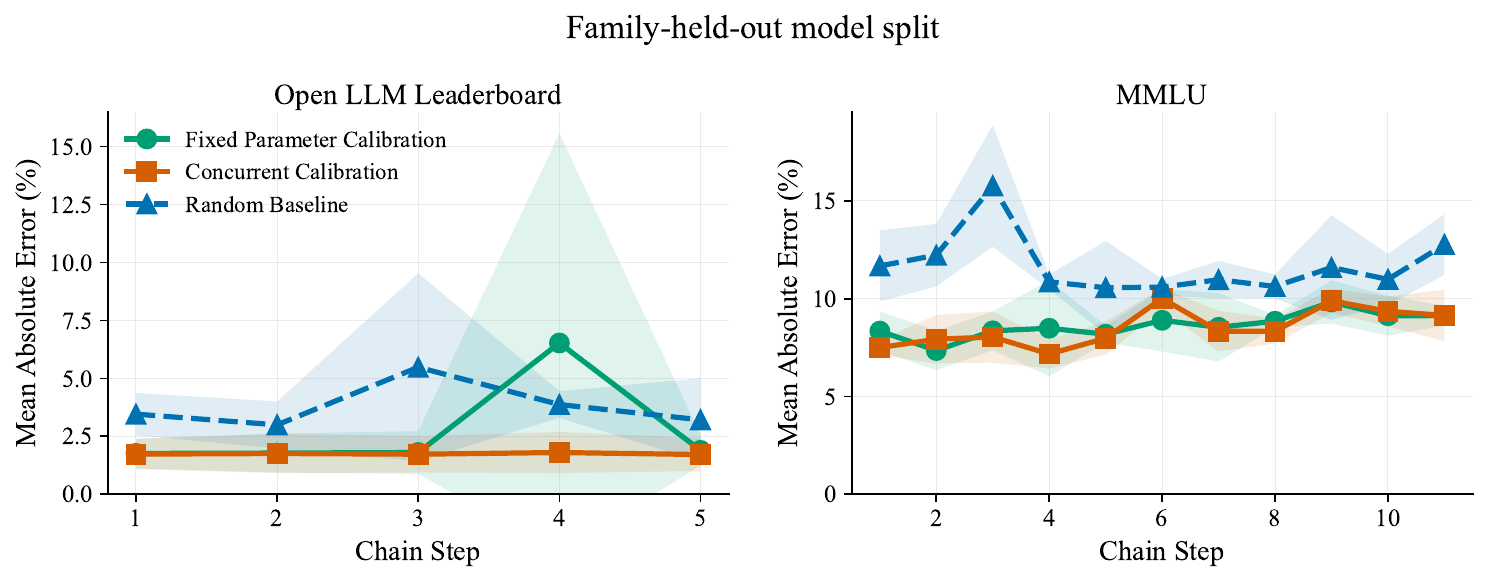}
    \caption{Prediction error remains low under time-ordered (top) and family-held-out (bottom) model splits, on the Open LLM Leaderboard (left) and MMLU (right). Fixed parameter calibration tracks concurrent calibration and outperforms random sampling, matching the random-split result in Figure~\ref{fig:anchor_ablation}. Shaded regions denote 95\% confidence intervals.}
    \label{fig:model_splits_appendix}
\end{figure*}

Figure~\ref{fig:model_splits_appendix} reports the full results under the time-ordered and family-held-out model splits on both benchmark suites. In the time-ordered split, models with earlier Open LLM Leaderboard submission dates are used as reference models and later models are held out for testing. In the family-held-out split, Llama and Mistral models are held out from the reference set. Across both the Open LLM Leaderboard and MMLU, fixed parameter calibration closely tracks concurrent calibration and generally outperforms random sampling, with prediction error remaining stable across chain steps.

\section{Implementation details}\label{app:impl}
We implement the MIRT 2PL model in py-irt with five latent dimensions. Parameters are estimated using stochastic variational inference with Adam (learning rate 0.1, exponential decay 0.9999) and the evidence lower bound (ELBO), with hierarchical priors on the item parameters. The initial calibration at $t=0$ and the concurrent  calibration baseline each run for 2000 epochs. Each fixed parameter calibration step runs for 1000 epochs, while holding the previously calibrated anchor parameters fixed. For anchor selection, we use K-means with Euclidean distance over the IRT item representations, with the number of clusters set to the anchor budget.

\end{document}